\documentclass[letter, 10 pt, conference]{ieeeconf}  

\IEEEoverridecommandlockouts                              

\usepackage[inter-unit-product =\cdot]{siunitx}
\usepackage{textcomp}
\usepackage{multirow}
\usepackage{subcaption}

\usepackage[pdftex]{graphicx}
\usepackage{cite}
\usepackage{amsmath}
\usepackage{amsfonts}
\usepackage{array}
\usepackage{url}
\usepackage{cleveref}
\usepackage{gensymb}
\usepackage{booktabs}

\usepackage{algorithm}
\usepackage{algorithmic}

\usepackage{adjustbox}

\usepackage{float}

\usepackage{comment}	
\usepackage{siunitx}    
\usepackage{amssymb}    
\usepackage{xcolor}		

\title{\LARGE \bf
	Conv1D Energy-Aware Path Planner for Mobile Robots in Unstructured Environments}

\author{Marco Visca,  
	Arthur Bouton, Roger Powell, Yang Gao
	and Saber Fallah
	\thanks{Marco Visca and Saber Fallah are with the Connected and Autonomous Vehicles Lab (CAV Lab), University of Surrey, Guildford, GU2 7XH, UK. Email: \{m.visca, s.fallah\}@surrey.ac.uk}
	\thanks{Arthur Bouton and Yang Gao are with the Space Technology for Autonomous and Robotic Laboratory (STAR LAB), Surrey Space Centre, University of Surrey, Guildford, GU2 7XH, UK. Email: \{a.bouton, yang.gao\}@surrey.ac.uk}
	\thanks{Roger Powell is in the Cybernetics Group, Remote Applications in Challenging Environments, UK Atomic Energy Authority, Culham Science Centre, OX14 3DB.}
}

\makeatletter
\def\endthebibliography{%
	\def\@noitemerr{\@latex@warning{Empty `thebibliography' environment}}%
	\endlist
}
\makeatother

\begin{document}

	\maketitle
	\thispagestyle{empty}
	\pagestyle{empty}
	
	\newdimen\origiwspc
	\newdimen\origiwstr
	\origiwspc=\fontdimen2\font
	\origiwstr=\fontdimen3\font
	\begin{abstract}
		Driving energy consumption plays a major role in the navigation of mobile robots in challenging environments, especially if they are left to operate unattended under limited on-board power. This paper reports on first results of an energy-aware path planner, which can provide estimates of the driving energy consumption and energy recovery of a robot traversing complex uneven terrains. Energy is estimated over trajectories making use of a self-supervised learning approach, in which the robot autonomously learns how to correlate perceived terrain point clouds to energy consumption and recovery. A novel feature of the method is the use of 1D convolutional neural network to analyse the terrain sequentially in the same temporal order as it would be experienced by the robot when moving. The performance of the proposed approach is assessed in simulation over several digital terrain models collected from real natural scenarios, and is compared with a heuristic inclination-based energy model. We show evidence of the benefit of our method \fontdimen3\font=0.1em to increase the overall prediction  r2 score by \SI{66.8}{\percent} and to reduce the driving energy consumption over planned paths by \SI{5.5}{\percent}.
		
	\end{abstract}
	

	%

	\section{Introduction}
	\setlength{\textfloatsep}{5pt}
	\fontdimen2\font=2.8pt
	Autonomous path planning of mobile robots in unstructured environments is a crucial task in many sectors such as: rescue robots for disaster area, agriculture, nuclear plants, and space exploration. Among different factors, energy consumption can play a major role on the success and efficiency of robotic missions. For example, planetary exploration rovers are required to drive several kilometres to reach potentially interesting scientific goals, but they often have limited on-board power resources  \cite{DBLP:journals/corr/abs-1805-05451}. Monitoring robots inside industrial plants or rescue robots in disaster areas are occasionally disconnected from umbilical cables and can be difficult to reach by human operators \cite{fukushima}. Estimating driving energy in advance can be vital to their operational safety and to allow for coverage of longer distances.
	
	In this work, we analyse the effect of unstructured geometries on driving energy consumption. While previous works have developed heuristic energy models that consider the terrain inclination as the main energy-relevant geometric factor \cite{9147470}\cite{7061469}\cite{6425519}\cite{7921087}\cite{Otsu2016}, their prediction performance can decrease in highly uneven surfaces. For example, scattered rocks, steps, bumps and rough terrain can induce complex motion dynamics which pose a greater challenge to the robot locomotion and control system, thereby leading to increased driving energy requirements. However, the complex interplay coming into action between wheel and terrain over unstructured geometries, and their effect on driving energy, can be challenging to model based on first principles. In this context, deep neural networks (DNNs) can be a well-suited asset, as they do not require explicit domain knowledge to be implemented into the prediction algorithm, and for their ability of extracting relevant features from high-dimensional data \cite{lecun2015deeplearning}. 
	
	In this paper,
	\begin{itemize}
		\fontdimen2\font=2.3pt
		\item We propose a DNN architecture to estimate the driving energy consumption and recovery of a mobile robot traversing unstructured terrains (Sections \ref{sub:data_preprocessing} and \ref{sub:nn_architeture}). The main novelty of this paper is the use of a Convolutional 1D (Conv1D) neural network to analyse the 3D geometry of the terrain from perceived point cloud data. In our approach, point clouds are rearranged in the same temporal order as they would be experienced by the robot when traversing the terrain, and a Conv1D neural network is trained to capture the sequential context of the wheel-terrain interaction from which driving energy depends.
		\fontdimen2\font=2.8pt
		\item The Conv1D neural network energy estimator is integrated into a lattice space A* path planner (Section \ref{sub:planning}). These two approaches are mutually beneficial to each other in that both energy estimation and path planning are achieved directly over feasible trajectories and unordered point clouds, thereby considering the actual traverse dynamic and the robot mobility constraints, and without any need of generating artificial 2D cost maps. 
		\item The robustness of this method is demonstrated over data from several real unstructured scenarios (Section \ref{sub:simulator}) and compared with a heuristic model which only considers the terrain inclination as geometric energy-relevant feature \cite{9147470}. Particularly, we show that our model is robust to increasing levels of terrain roughness, while the latter substantially degrades its performance as the terrains deviate from planar (Sections \ref{sub:test_results} and \ref{sub:planning_results}). Finally, we provide evidence of its benefit to reduce driving energy consumption (Section \ref{sub:planning_results}).
	\end{itemize}
	
	To the best of the authors knowledge, this work is the first to propose an integrated energy-aware prediction and planning framework for mobile robots which considers the effect of complex unstructured terrain geometries.

	\section{Related Work}\label{sec:related_works}
	\fontdimen2\font=3pt
	Existing algorithms in the field of path planning in unstructured environments operate on a 2D-grid superimposed over the operational map \cite{surrey721940}\cite{GESTALT}. Each cell of the grid is assessed with specific traversibility tests and, based on their results, assigned with a cost value. While this could be an effective and simple method to ensure reliable and safe navigation, expressing energy cost in this form can be challenging, as it assumes that each cell has a well defined intrinsic cost value. Besides, the cost functions used by grid methods are usually heuristic weighted combinations of terrain geometric properties such as slope, and roughness \cite{4161272}\cite{exomarsgnc}\cite{6094768}. However, they do not provide an explicit estimation of energy consumption.


	Other works have proposed the use of accurate onboard traverse simulators, which take as input geometric and other terrain information, and explicitly run a forward navigation simulation so as to assess the energy consumption of specific trajectories \cite{6094768}\cite{JPL_simulator}\cite{doi:10.1002/rob.21483}. However, in spite of their accuracy, their computational workload is often excessive for real-time navigation, making them cumbersome to integrate into a path planning optimization framework.

	Several machine learning techniques have been proposed to improve the path planning autonomy of mobile robots. This includes terrain classification \cite{survey11}\cite{JPL_classification}\cite{rlpaper17}, obstacle detection \cite{gao_seeker}, slip prediction \cite{doi:10.1002/rob.21761}\cite{DBLP:journals/corr/abs-1806-07379}, and terramechanical parameters estimation \cite{terramec1}\cite{gaussian} among the others. Specifically, artificial neural networks have gained an increasing popularity in this context for their ability to extract features from high-dimensional inputs, and their efficient parallel computing \cite{lecun2015deeplearning}\cite{simone}. Similarly to our work, Higa et al. \cite{8764007} have proposed a deep learning method to estimate driving energy of a mobile robot on uneven terrains. Their implementation has the advantage to consider not only terrain geometry, but also its visual aspect by using a Convolutional 2D (Conv2D) neural network. However, in their work driving energy is correlated over squared patches of terrain. This can be considered as a static approximation, which associates an energy value to a terrain patch, but which does not capture the sequential context over the actual trajectory. Furthermore, they do not show how their method can be integrated into a path planner.

	
	\begin{figure}[t]
		\centering
		\begin{subfigure}[b]{0.25\linewidth}
			\includegraphics[width=\textwidth]{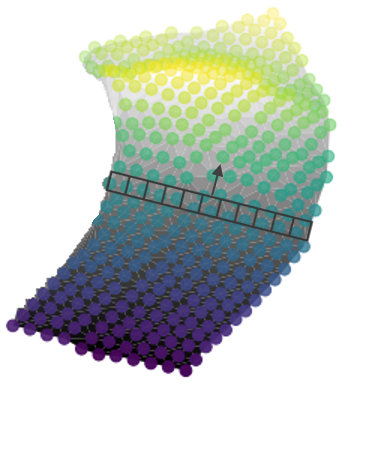}
			\caption{}
			\label{fig:point_clouds_processing_points}
		\end{subfigure}
		\begin{subfigure}[b]{0.24\linewidth}
			\includegraphics[width=\textwidth]{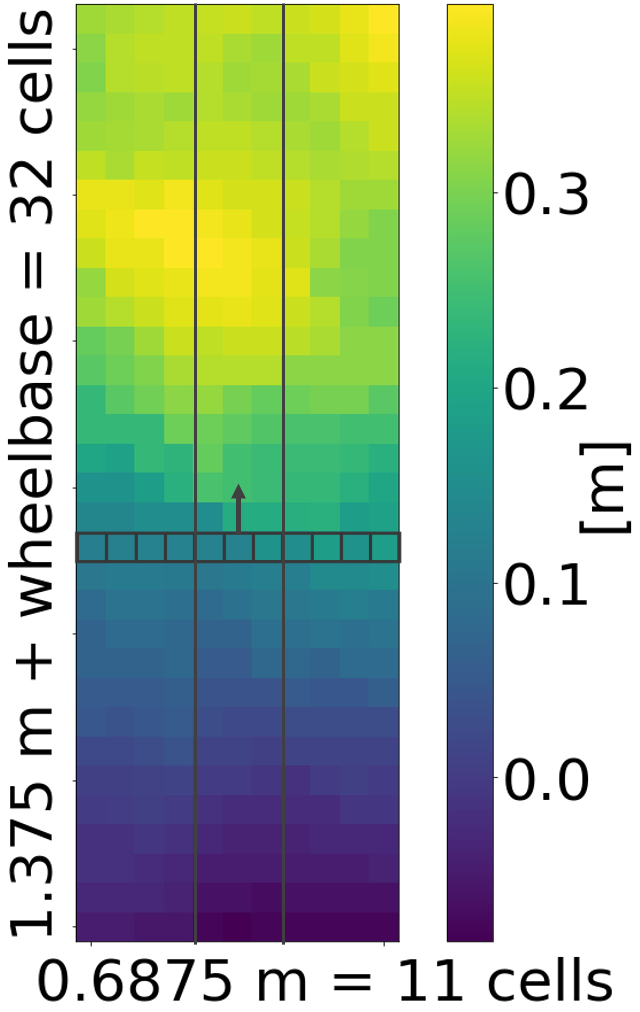}
			\caption{}
			\label{fig:point_clouds_processing_trace_tot}
		\end{subfigure}
		\begin{subfigure}[b]{0.22\linewidth}
			\includegraphics[width=\textwidth]{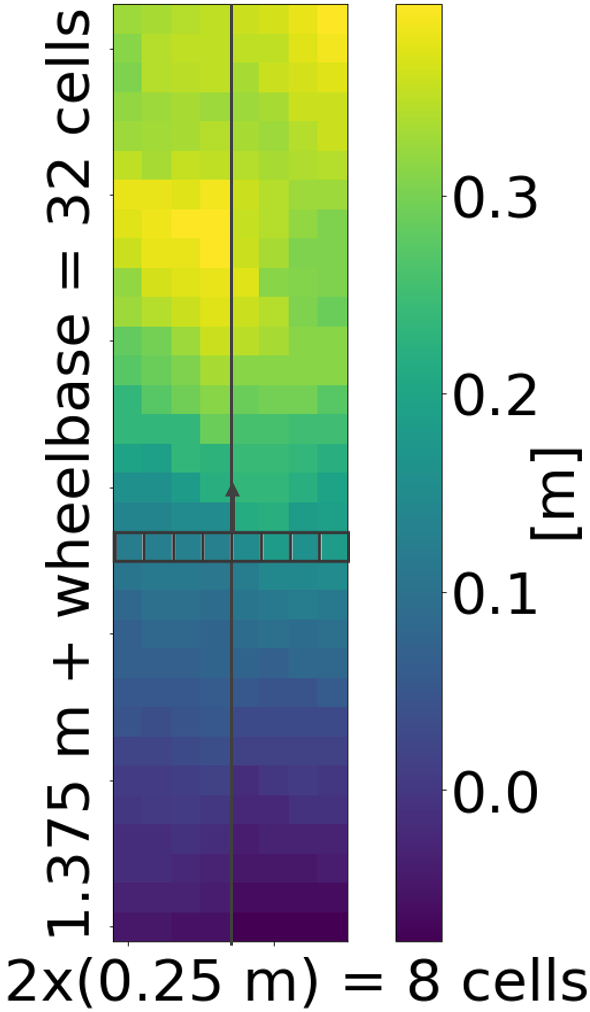}
			\caption{}
			\label{fig:point_clouds_processing_trace_final}
		\end{subfigure}
		
		\caption{Point Cloud data preprocessing pipeline. (a) Point cloud relative to trajectory is extracted from memory. (b) Point cloud is downsampled into a 2D voxel-grid. (c) The central region not in contact with the wheels is removed. }
		\label{fig:point_clouds_processing}
	\end{figure}

	Different works have developed energy-aware path planners which explicitly estimates energy over feasible trajectories \cite{9147470}\cite{7061469}\cite{6425519}\cite{7921087}. However, they make use of heuristic energy models that only consider the terrain inclination as geometric energy-relevant feature. We argue that, while these approaches can be satisfactory for relatively planar surfaces, they cannot capture the influence of more complex unstructured geometries. A similar implementation of the energy model developed in Gruning et al. \cite{9147470}, valid for our robotic platform, is used as main reference for qualitative and quantitative comparisons. We refer to it as RampModel in the rest of the paper. Furthermore, in their work two non-geometric factors are additionally considered for energy estimation: surface friction, and internal friction due to the track speed. However, as this work focuses on the effect of unstructured terrain geometries, surface friction and track speed are maintained constant during our experiments.

	\section{Constructing the Energy-Aware Path Planner}\label{sec:method}
	\fontdimen2\font=2.6pt
	We propose to estimate driving energy end-to-end from point clouds by neural network training. Particularly, the neural network must be able to analyse the sequential context of the traversed terrain directly over feasible trajectories. In this way, driving energy estimation can be achieved by considering the actual traverse dynamics and the robot mobility constraints, thereby increasing the prediction accuracy. Moreover, the proposed approach allows to plan directly over unordered point clouds, without any need of cost map generation. This can be advantageous as point clouds can be easily obtained from any kind of range sensor, they are easy to maintain, and they can be split and merged with minimal computational effort \cite{doi:10.1002/rob.21700}. The remainder of this section illustrates the proposed methodology.
	
	
	\subsection{Point Cloud Collection and Preprocessing} \label{sub:data_preprocessing}\label{sub:preprocessing}
	A diagram  of the point cloud collection and preprocessing procedure is illustrated in Fig. \ref{fig:point_clouds_processing}. Points are retrieved from memory along the trajectory with a width equal to the wheel track (Fig. \ref{fig:point_clouds_processing_points}).
	A safety check is performed before continuing with the preprocessing pipeline. The point cloud is analysed with standard geometric features extraction techniques to measure pitch, roll, and residual features \cite{surrey721940}. If at least one of these values is above specific thresholds the trajectory is labelled as untraversable and is not considered for the energy planning optimisation. Conversely, if the point cloud passes the safety check, the preprocessing procedure can be continued. The point cloud is downsampled with a 2D-voxel grid having each of its row perpendicular to the local direction of the trajectory (Figs \ref{fig:point_clouds_processing_points} and \ref{fig:point_clouds_processing_trace_tot}). In this way, each row can be seen as the elevation points seen under a section of the rover along its width at a specific time, while each column represents the evolution of the terrain features under a specific location of the rover as it advances along the trajectory. Downsampling is motivated by the need of the neural network to have a fixed number of inputs to be used. Each cell of the voxel-grid is filled by averaging the elevation of all points in that cell, and by linear interpolation if no point is present. This choice does not introduce notable artefacts as long as the point cloud has a similar or higher density than the voxel discretization \cite{1087212}. For our application, and in line with current on-board perception technologies \cite{surrey812938}, a uniform point cloud density of \SI{256}{pts/m^2} is assumed in a region up to \SI{4}{\meter} distant from the robot. Next, the voxel discretization is set to a value of \SI{6.25}{cm} $\times$ \SI{6.25}{cm}. Finally, we can observe that all the elevation points in the central region of the voxel-grid have lower elevations than the rover belly, as they have passed the safety check. As a result, they do not interact with the robot and thereby do not impact the driving energy. Hence, only the two lateral bands, of width \SI{25}{cm} each, are retained (Fig. \ref{fig:point_clouds_processing_trace_final}). \fontdimen2\font=3.1pt The result of the point cloud preprocessing analysis is a \num{32} $\times$ \num{8} matrix of elevation values, which represents one input to the proposed neural network. \fontdimen2\font=\origiwspc
	
	
	\begin{figure}[t]
		\centering
		\includegraphics[scale=0.2]{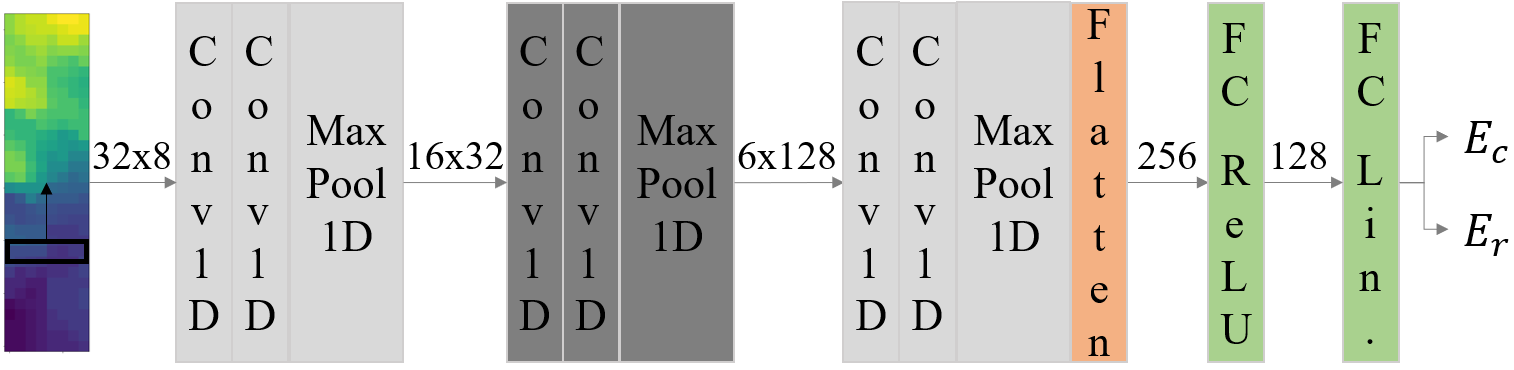}
		\caption{Conv1D neural network architecture.}
		\label{fig:Conv1d}
	\end{figure}

	\subsection{Neural Network Architecture}\label{sub:nn_architeture}
	We propose to learn driving energy from point clouds by means of 1D convolutional neural network.
	Conv1D is a well known neural network architecture which has demonstrated remarkable performance in analysis of series of sensor data, signal processing and natural language processing \cite{kiranyaz20191d}. Additionally, it has the advantage over alternative sequential methods (e.g. recurrent neural networks) to require lower training effort and to better exploit parallel computing \cite{peddinti2015time}. Similarly, we propose to formulate our problem in a sequential fashion. Hence, Conv1D can be used to analyse the terrain sequence of Fig. \ref{fig:point_clouds_processing_trace_final} by means of a fixed-dimension 1D context window over the temporal (i.e. vertical) axis.
	
	Fig. \ref{fig:Conv1d} shows the proposed Conv1D architecture. It consists of three stacks of two 1D convolutional layers followed by 1D max pooling. In this way, the bottom layers have the ability to learn a narrow temporal context, while wider temporal relationships can be learned at deeper layers. The first stack has kernel dimension $K=2$ and zero padding, while the second and third stacks have $K=3$ and no padding. The output of Conv1D is flattened and processed by one Fully Connected (FC) layer with \num{128} units and ReLU activation, followed by a FC layer with \num{2} units and linear activation. The outputs are the predicted energy consumption $E_c$ and energy recovery $E_r$.

	\subsection{Planning Algorithm}\label{sub:planning}
\begin{figure}[t]
		\centering
		\begin{subfigure}[b]{0.24\linewidth}
			\includegraphics[width=\textwidth]{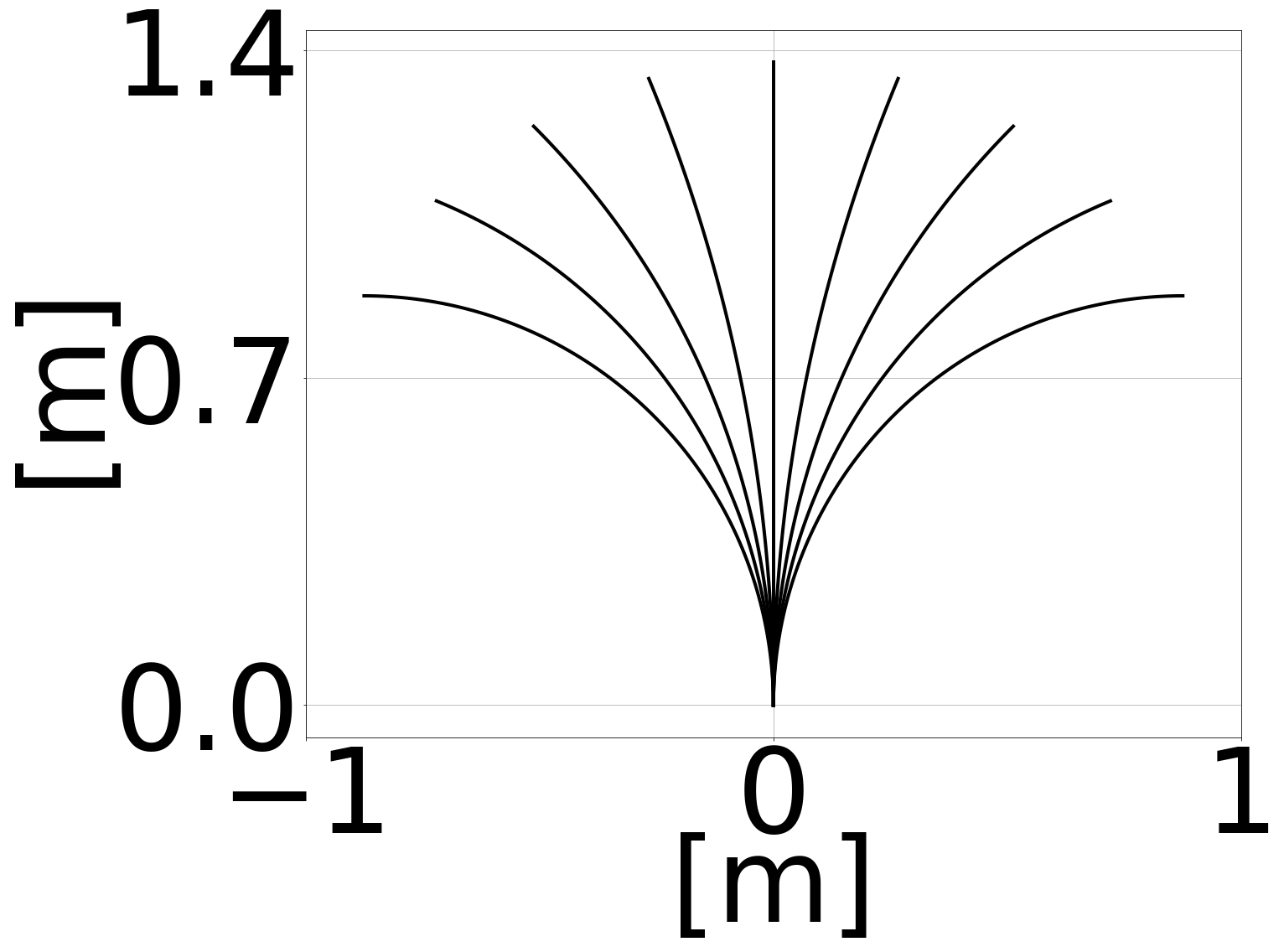}
			\caption{}
			\label{fig:action_space}
		\end{subfigure}
		\begin{subfigure}[b]{0.725\linewidth}
			\includegraphics[width=\textwidth]{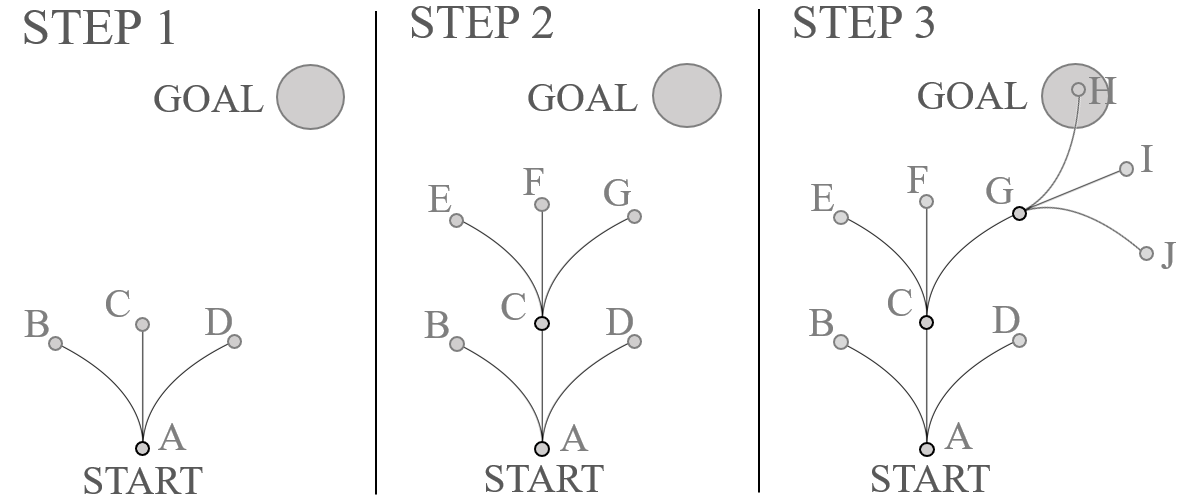}
			\caption{}
			\label{fig:a_star}
		\end{subfigure}
		
		\caption{(a) The 9 elementary trajectories of the lattice space, and (b) diagram of A* optimisation in lattice state space with only three actions for simplicity.}
		\label{fig:path_planning}
	\end{figure}

	Motion planning is performed in a state lattice space, a well known approach to the problem of differentially constrained mobile robot planning in unstructured environments \cite{pivtoraiko2009differentially}. State lattice is a search graph where vertices representing kinematic states of the robot are connected by edges representing trajectories that satisfy its kinematic constraints. In this way, planning and cost estimation can be achieved directly over feasible trajectories. In our application, we define a set of \num{9} elementary trajectories \SI{1.375}{m} long, and with curvature uniformly spaced in  [\num{-1.14} , \num{1.14}] \SI{}{m^{-1}}, \fontdimen2\font=2.3pt according to the mobility capability of our robot (see Fig. \ref{fig:action_space}). \fontdimen2\font=\origiwspc
	
	The A* graph search optimisation is then used in the lattice space. This choice is motivated by the proven optimality of A* given a consistent heuristic \cite{a_star_opt}, and by its simplicity and effectiveness to optimise paths according to complex cost functions, making it well-suited to the robotic navigation in unstructured terrains \cite{exomarsgnc}. Fig. \ref{fig:a_star} shows a diagram of the A* planning process for a lattice space with three actions. In the diagram, the start position is labelled as node A and added to a list called OPEN, which contains all nodes that have been found but not yet expanded. At each iteration of A*, a new node in the OPEN list is chosen according to its priority and expanded for the set of all elementary trajectories. In A* the priority is expressed as:
	\vspace{-2.5mm}
	
	\begin{equation}\label{eq:priority}
	P(n) = C(n_p,n) + H(n,g)
	\end{equation}
	
	\vspace{-0.mm}
	The priority $P$ of node $n$ is given by the real cost $C(n_p,n)$ of moving from the parent node $n_p$ to $n$, and the heuristic estimation $H(n,g)$ of the remaining cost from $n$ to the goal $g$. The new node is then chosen as the one having the lowest $P$. The search continues until a goal node is retrieved from the OPEN list. At this point, by traversing backward the stored parent-child relationships, the path and the associated cost of A-C-G-H is found from the start to the goal. For more details on A* refer to \cite{astar}.
	
	In our application, $C(n_p,n)$ and $H(n,g)$ must have a meaning of driving energy. Particularly, the term $C(n_p,n)$ must represent the true value of energy when moving from state $n_p$ to $n$. We propose a novel method to provide an estimate to this problem. For each new trajectory in the graph, the relative point cloud is retrieved from memory and preprocessed according to \ref{sub:data_preprocessing} and fed to the Conv1D neural network described in \ref{sub:nn_architeture}. Its transition cost is then defined as: $C(n_p,n) = E_c-E_r$. On the contrary, the heuristic $H(n,g)$ does not require to provide the true energy value, but must be sufficiently informative to address the search algorithm in promising directions. The same $H(n,g)$ of the RampModel implementation is used, which assumes that the goal can be reached by moving on a straight line, and the heuristic for remaining energy is defined as:
	\vspace{-0.5mm}
	\begin{equation}\label{eq:heuristic}
	H(n,g) = (G\theta + \beta)d
	\end{equation}
	\vspace{-4.5mm}

	Where $\theta$ and $d$ are respectively the relative inclination in degree and \fontdimen3\font=0.1em euclidean distance in meters between $n$ and $g$, while $G$ and $\beta$ are heuristic parameters set to provide optimistic energy estimates. In this way, the consistency of the heuristic is guaranteed, thereby enabling the convergence of A* to an optimal solution. For our application, $\beta$ is set to \SI{1.90}{J/m}, while $G$ is set to \SI{6.13}{J/m} if $\theta$ is positive (i.e. uphill), and \SI{1.18}{J/m} if $\theta$ is negative (i.e. downhill).
	

	\begin{figure}[t]
		\centering
		\begin{subfigure}{0.24\linewidth}
			\includegraphics[width=\textwidth]{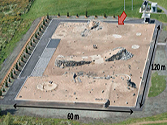}
			\caption{}
			\label{fig:met}
		\end{subfigure}
		\begin{subfigure}{0.51\linewidth}
			\includegraphics[width=\textwidth]{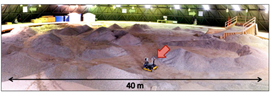}
			\caption{}
			\label{fig:utias}
		\end{subfigure}
		
		\begin{subfigure}{0.24\linewidth}
			\includegraphics[width=\textwidth]{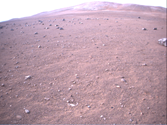}
			\caption{}
			\label{fig:seeker}
		\end{subfigure}
		\begin{subfigure}{0.24\linewidth}
			\includegraphics[width=\textwidth]{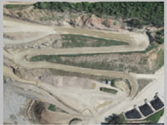}
			\caption{}
			\label{fig:quarry}
		\end{subfigure}
		\begin{subfigure}{0.24\linewidth}
			\includegraphics[width=\textwidth]{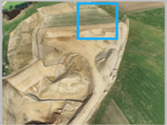}
			\caption{}
			\label{fig:gravelpit}
		\end{subfigure}
		
		\caption{ Training dataset is (a) Canadian Space Agency (CSA) Martian Emulation Terrain (MET). Test datasets are (b) UTIAS indoor rover test facility, (c) Atacama desert SEEKER test trial, (d) mining quarry, and (e) gravel pit.}
		\label{fig:real_envs_images}
	\end{figure}
	
	\begin{figure}[t]
		\centering
		\begin{subfigure}[b]{0.39\linewidth}
			\includegraphics[width=\textwidth]{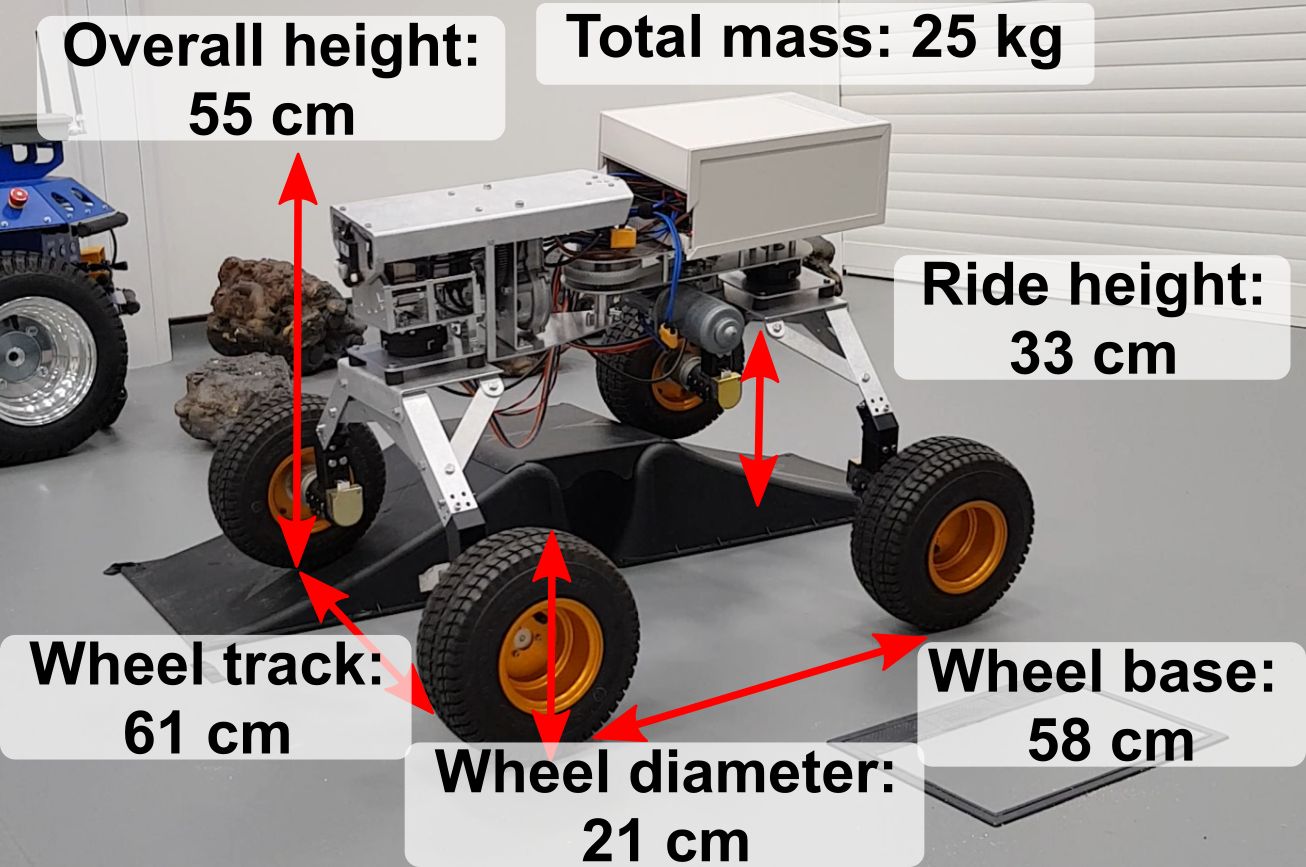}
			\caption{}
		\end{subfigure}
		\begin{subfigure}[b]{0.294\linewidth}
			\includegraphics[width=\textwidth]{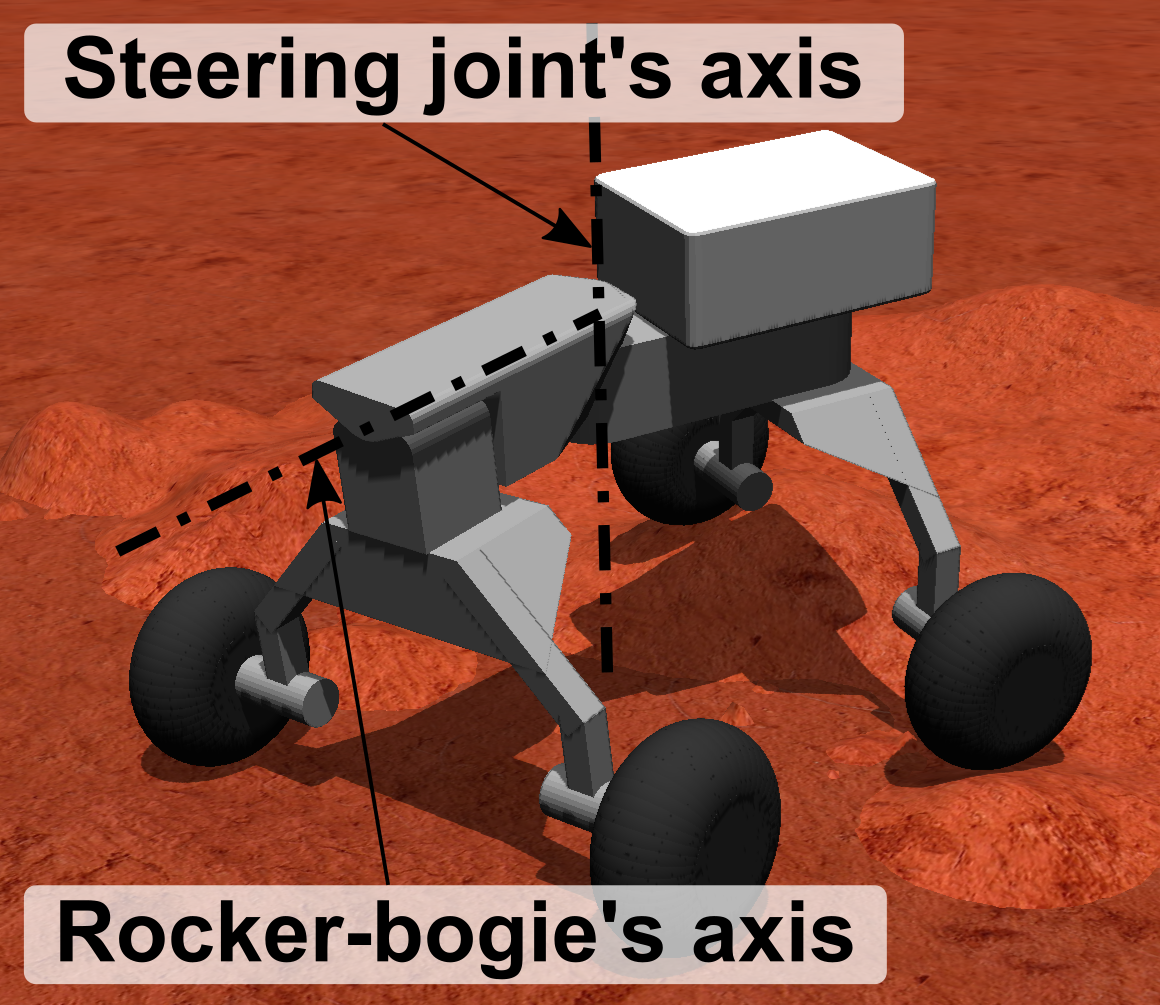}
			\caption{}
		\end{subfigure}
		\begin{subfigure}[b]{0.29\linewidth}
			\includegraphics[width=\textwidth]{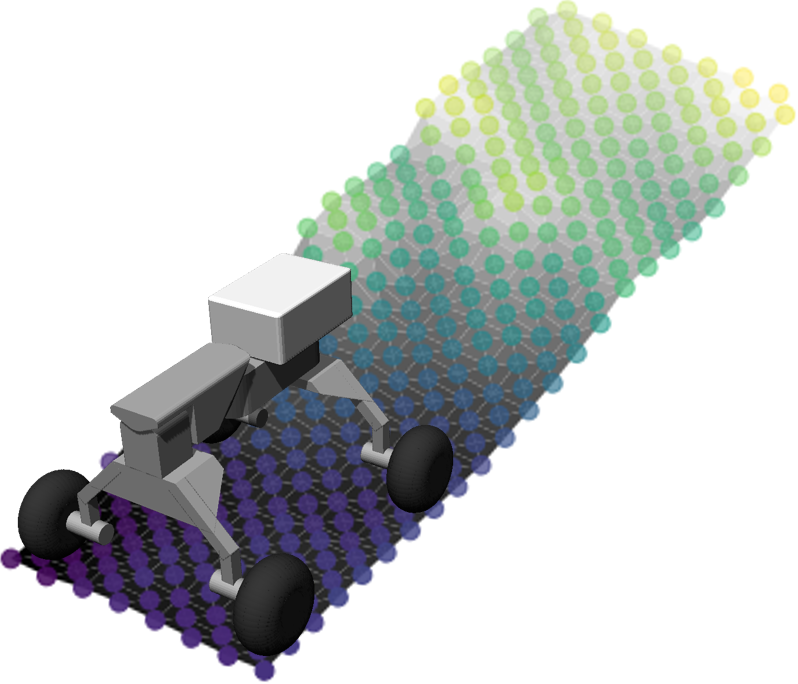}
			\caption{}
			\label{fig:terrain}
		\end{subfigure}
		\caption {(a) Features of the rover MARCEL. (b) Simulation model of the rover and its mobilities. (c) Example of the rover facing a trajectory sample to be analysed.}
		\label{fig:robot}
	\end{figure}


%
%

	\section{Experimental Setup}\label{sec:simulating_assets}
	\fontdimen2\font=2.6pt
	\subsection{Simulation Environments}\label{sub:simulator}
	The algorithm is developed and tested using a physical simulator implemented in C++ and based on Open Dynamics Engine (ODE) \cite{7989281}\cite{7487448}. Different simulation environments are integrated in ODE from data collected from real uneven terrains (Fig. \ref{fig:real_envs_images}). Three of them are from Martian-like scenarios and collected using laser scans on-board wheel robotic platforms (Figs \ref{fig:met}, \ref{fig:utias} and \ref{fig:seeker}). The other two are from a mining quarry and a gravel pit scenarios and collected with a fixed-wing flying robot using structure-from-motion techniques (Figs \ref{fig:quarry} and \ref{fig:gravelpit}). The datasets are all available online at \cite{asrl} and \cite{sensefly} with the exception of the SEEKER dataset \cite{seeker}.
	
	\subsection{Robot Model}\label{sub:robot}
	\fontdimen2\font=3pt
	The numerical model of the robot is based on the features of the actual four-wheels rover prototype MARCEL first presented in \cite{marcel} and depicted in Fig. \ref{fig:robot}. The rover is fitted with a rocker-bogie at the rear which ensures the contact of all wheels with the ground at all times. Indeed, this can be critical to guarantee consistent and energy-efficient driving over unstructured geometries. The rover is also provided with an actuated vertical pivot at the centre that allows it to steer smoothly. The wheels are synchronized with the steering joint rotation so that the rover can maneuver with theoretically no tangential and lateral skidding on a flat surface. The wheels are powered by four independent DC-motors subjected to PI speed controllers, while the steering joint has its own DC-motor and PID angular position controller. All the controllers run at a frequency of 1 kHz.
	
	\fontdimen2\font=2.6pt
	\subsection{Ground Truth Energy Collection}\label{sub:energy}
	This section illustrates how ground truth energy data are collected for training of the network described in Section \ref{sub:nn_architeture}. The power of each motor is computed independently using the steady state equivalent circuit of the DC-motor armature \cite{8045}. The power of a DC motor is given by:
	\vspace{-2mm}
	
	\begin{equation}\label{eq:power1}
	P = \frac{\omega\tau}{\eta} + R_aI_a^2   
	\end{equation}
	
	
	The first term is the mechanical power, where $\omega$ is the motor angular speed, $\tau$ is the torque to the motor, and $\eta$ is the motor efficiency due to mechanical losses. Mechanical power can be either consumed, if $\omega$ and $\tau$ have the same sign, and recovered, if they have opposite sign. The second term is the electrical power loss due to the motor resistance dissipated as heat, where $R_a$ is the armature resistance and $I_a$ is the current flowing into the motor. The current can also be expressed as follows:
	\vspace{-1.mm}
	\begin{equation}\label{eq:current}
	I_a=\frac{\tau} {K_T g_r \eta} 
	\end{equation}
	
	Where, $K_T$ is the motor torque constant, and $g_r$ is the gear ratio. It follows that the power can be expressed as a function of $\omega$ and $\tau$ as:
	\vspace{-1mm}
	\begin{equation}\label{eq:power2}
	P =\frac{\omega\tau}{\eta} + \left(\frac{\tau} {K_T g_r \eta} \right)^2R_a
	\end{equation}
	
	Simulated sensors on-board the robot measure $\omega$ and $\tau$ for each motor with a constant sampling rate $\Delta t$, and the power is computed with the formula in \ref{eq:power2}. Finally, assuming constant power between each $\Delta t$, the energy consumption $E_c$ and the energy recovery $E_r$ over the time interval $[t_0, t_f]$ can be computed as:
	\vspace{-1mm}
	\begin{equation}\label{eq:energy}
	\begin{cases}
	E_c =  \sum\limits_{t_0}^{t_f}P_t{\Delta t} & P_t \geq 0 \\
	E_r = \sum\limits_{t_0}^{t_f}P_t{\Delta t} & P_t < 0
	\end{cases}       
	\end{equation}
	
	For our platform, we assume: $\eta = 0.83$  if $\omega$ and $\tau$ have the same sign, and $\eta = 3.33$ if they have opposite sign, $R_a = $ \SI{0.5}{\ohm}, $K_T = $ \SI{0.05}{\newton\meter\per\ampere}, $g_r = 62$, and $\Delta t = 10^{-3}s$.
	

	\begin{table*}[t]
		\centering
		\caption{Energy Estimation Test Results.}
		\label{tab:test_results}
		\begin{tabular}{llcccccccccccccc}
			\multicolumn{2}{c}{Dataset} & &Model&& \multicolumn{3}{c}{Consumption} && \multicolumn{3}{c}{Recovery} && \multicolumn{3}{c}{Total}\\
			
			\cmidrule{1-2} \cmidrule{6-8} \cmidrule{10-12} \cmidrule{14-16}
			
			Name & Samples &&  && MAE & MSE & R2 && MAE & MSE & R2 && MAE & MSE & R2\\
			
			Ramps & 1881 && Conv1D && 5.353 & 68.445 & 0.971 && 0.947 & 1.746 & 0.982 && 5.354 & 68.564 & 0.976 \\
			&  && RampModel && 2.112 & 10.231 & \textbf{0.996} && 0.434 & 0.360 & \textbf{0.996} && 2.078 & 9.812 &\textbf{0.996} \\
			&  && Conv1D (\SI{25}{\percent}) && 7.074 & 93.293 & 0.961 && 1.387 & 3.202 & 0.968 && 6.899 & 91.212 & 0.968 \\[0.05cm]
			
			MET (val) & 21578 && Conv1D && 4.804 & 65.526 & \textbf{0.939} && 0.568 & 0.654 & \textbf{0.983} && 4.549 & 60.792 & \textbf{0.956} \\
			&  && RampModel && 14.182 & 385.477 & 0.640 && 1.169 & 2.743 & 0.930 && 13.343 & 342.075 & 0.750 \\
			&  && Conv1D (\SI{25}{\percent}) && 5.505 & 78.255 & 0.927 && 0.687 & 0.906 & 0.977 && 5.181 & 72.217 & 0.947 \\[0.05cm]
			
			SEEKER & 33673 && Conv1D && 3.024 & 16.938 & \textbf{0.973} && 0.509 & 0.456 & \textbf{0.982} && 2.755 & 14.234 & \textbf{0.982} \\
			&  && RampModel && 10.361 & 153.691 & 0.751 && 1.324 & 3.044 & 0.880 && 9.185 & 119.125 & 0.847 \\
			&  && Conv1D (\SI{25}{\percent}) && 3.626 & 22.400 & 0.964 && 0.671 & 0.754 & 0.970 && 3.278 & 18.556 & 0.976 \\[0.05cm]
			
			UTIAS & 11668 && Conv1D && 6.656 & 98.209 & \textbf{0.906} && 1.010 & 1.910 & \textbf{0.903} && 6.172 & 89.859 & \textbf{0.919} \\
			&  && RampModel && 21.969 & 717.477 & 0.315 && 2.843 & 11.393 & 0.421 && 19.423 & 588.190 & 0.469 \\
			&  && Conv1D (\SI{25}{\percent}) && 7.599 & 120.944 & 0.885 && 1.327 & 3.341 & 0.830 && 6.885 & 105.803 & 0.904 \\[0.05cm]
			
			Gravelpit & 790 && Conv1D && 4.023 & 43.906 & \textbf{0.922} && 0.731 & 1.082 & \textbf{0.925} && 3.565 & 36.843 & \textbf{0.931} \\
			&  && RampModel && 12.450 & 321.789 & 0.427 && 1.604 & 4.751 & 0.670 && 10.883 & 256.604 & 0.517 \\
			&  && Conv1D (\SI{25}{\percent}) && 4.761 & 54.772 & 0.902 && 1.182 & 2.159 & 0.850 && 3.963 & 42.913 & 0.919 \\[0.05cm]
			
			Quarry & 4699 && Conv1D && 7.497 & 144.139 & \textbf{0.851} && 1.120 & 2.517 & \textbf{0.923} && 6.968 & 133.121 & \textbf{0.865} \\
			&  && RampModel && 21.585 & 817.665 & 0.156 && 2.686 & 11.772 & 0.639 && 19.181 & 680.424 & 0.308 \\
			&  && Conv1D (\SI{25}{\percent}) && 8.550 & 172.919 & 0.822 && 1.567 & 4.140 & 0.873 && 7.706 & 153.793 & 0.844 \\
			
			%
			
		\end{tabular}
	\end{table*}
	
	\begin{table*}[t]
		\centering
		\setlength\belowcaptionskip{-3.5pt}
		\caption{Path Planner Test Results.}
		\label{tab:path_planner_test_results}
		\begin{tabular}{llccccccccccc}
			&  && \multicolumn{3}{c}{Planning Performance} && \multicolumn{3}{c}{Prediction Performance} && \multicolumn{2}{c}{Total Driving Energy} \\
			
			\cmidrule{4-6} \cmidrule{8-10} \cmidrule{12-13}
			
			Targets & Model && Nodes & Avg. Node Time [s] & Tot Time [s] && MSE & MAE & R2 && Predicted [J] & Real [J]\\
			
			259 & Conv1D && 5619 & 0.212 & 1265 && 271.938 & 7.882 & 0.937 && 16.268k & 17.108k\\
			& RampModel && 2769 & 0.208 & 590 && 2514.917 & 33.724 & 0.269 && 9.368k & 18.103k\\
			& DynSim && 3692 & 5.586 & 20699 && 0 & 0 & 1 && 15.623k & 15.623k\\[0.05cm]
			
			1 (Fig. \ref{fig:path_planning_example}) & Conv1D && 16 & 0.183 & 3.012 && - & - & - && 29.23 & 30.44\\
			& RampModel && 4 & 0.182 & 0.808 && - & - & - && -0.99 & 39.33\\
			& DynSim && 12 & 4.324 & 51.750 && - & - & - && 27.72 & 27.72\\
		\end{tabular}
		
		\vspace{-4.1mm}
	\end{table*}

	\subsection{Training Dataset}\label{sub:training_dataset}
	We train the network by collecting data from the Canadian Space Agency (CSA) Martian Emulation Terrain (MET) (Fig. \ref{fig:met}), while we use the other environments for testing. The robot traverses the \textit{MET} environment at constant speed of \SI{20}{cm/s} and collects data from \num{107606} trajectories, equivalent to approximately \SI{150}{km} of traverse. The dataset is then randomly divided between training (\SI{80}{\percent}) and validation (\SI{20}{\percent}) datasets. Finally, the network is trained by means of stochastic gradient descent for \num{100} epochs (about 3 hours on a GeForce RTX 2080 Ti GPU) with mean squared error loss, \num{64} mini batch size, \num{e-4} learning rate, and RMSprop optimizer \cite{rmsprop}.
	
	\section{Results}\label{sec:results}
	\fontdimen2\font=2.6pt
	\subsection{Test Results}\label{sub:test_results}
	The performance of Conv1D and RampModel are summarised in Table \ref{tab:test_results}. The first dataset, called \textit{Ramps}, is a synthetic environment composed of perfectly planar ramps. The robot traverses these ramps with 1881 trajectories having different combinations of pitch, roll and curvature. Conv1D has slightly lower performance than RampModel on \textit{Ramps}. This can be explained by the fact that the former is a machine learning model which has been trained exclusively with data from unstructured environments, while the latter is a heuristic method ad-hoc devised for planar ramps. Nevertheless, both models retain satisfactory predictions, with Conv1D and RampModel total r2 score of \SI{97.6}{\percent} and \SI{99.6}{\percent} respectively. The results in the \textit{Ramps} dataset show that RampModel accurately predicts driving energy in planar scenarios. However, we observe that its accuracy significantly degrades in unstructured surfaces, while Conv1D retains considerably better performance. Particularly, the RampModel total r2 score decreases to values of \SI{30.8}{\percent} on \textit{Quarry}, \SI{46.9}{\percent} on \textit{UTIAS}, \SI{51.7}{\percent} on \textit{Gravelpit}, \SI{75.0}{\percent} on \textit{MET}, and \SI{84.7}{\percent} on \textit{SEEKER}, while Conv1D is able to retain values of \SI{86.5}{\percent}, \SI{91.9}{\percent}, \SI{93.1}{\percent}, \SI{95.6}{\percent}, and \SI{98.2}{\percent} respectively. We observe that the RampModel prediction of energy consumption is more largely affected than the energy recovery. A likely reason for this is that the energy recovery mostly depends on the downhill inclination of the terrain, making it less challenging, but still far from accurate, to estimate with RampModel. In contrast, the presence of steps, bumps, rough terrain and other complex geometries has a significant impact on the energy consumption, which can not be captured by an energy model exclusively based on terrain inclination. Conversely, Conv1D learns implicitly, by analysing the sequential context of the traversed terrain, the most relevant features to be correlated with driving energy and, in this way, retains better generalisation to the unforeseen unstructured geometries. As a qualitative example, we analyse the terrain illustrated in Fig. \ref{fig:terrain}. It consists of an uphill trajectory with a series of scattered bumps. RampModel predicts a total driving energy of \SI{54.99}{J}, which would be an accurate estimate for a planar terrain with the same inclination. However, the scattered bumps induce a greater energy demand on the robot actuators increasing the actual energy value to \SI{108.29}{J}. As with previous results, this effect is more accurately captured by Conv1D, which predicts a total driving energy of \SI{107.71}{J}.
	
	We also test the performance of Conv1D when the training samples are reduced to \SI{25}{\percent} of the original training dataset. A limited degradation of total r2 score can be observed with differences ranging from \SI{0.6}{\percent} to \SI{2.1}{\percent} depending on the test dataset. This shows evidence of the capability of Conv1D to retain similar performance with limited amount of data, which can be a crucial aspect in real-world applications.

	
	\subsection{Path Planner Integration Results}\label{sub:planning_results}
	Conv1D and RampModel energy estimators are integrated into the A* lattice path planner. Their performance are assessed by conducting statistical analysis over \num{259} random start-goal positions within the different test environments. Table \ref{tab:path_planner_test_results} summarises our findings. We observe that the number of nodes expanded by Conv1D is double that of RampModel. Specifically, over the \num{259} search problems Conv1D and RampModel expand respectively \num{5619} and \num{2769} nodes. A plausible explanation is that Conv1D has a larger disparity between the energy estimated by the heuristic $H(n,g)$ and the actual cost $C(n_p,n)$. An intrinsic property of the A* optimization is that the more optimistic the heuristic is compared to the actual cost, the more A* has to expand the search. Therefore, Conv1D has to analyse a higher number of possible trajectories prior to find the optimal solution. While this results in longer planning time, Conv1D is nevertheless able to provide solutions which require lower driving energy consumption, and with considerably higher prediction accuracy. Specifically, the optimal predicted and real driving energy according to Conv1D is of respectively \SI{16.268}{kJ} and \SI{17.108}{kJ}, with a r2 score of \SI{93.7}{\percent}. Conversely, RampModel provides a predicted solution of \SI{9.368}{kJ}, but which has an actual energy demand of \SI{18.103}{kJ} and a r2 score of \SI{26.9}{\percent}. Therefore, Conv1D is able to provide solutions which consume \SI{5.5}{\percent} less driving energy and to improve the r2 score prediction accuracy by \SI{66.8}{\percent}. We also compare the two approaches with a ground truth method, called DynSim. In this method, the dynamic simulator, described in Section \ref{sec:simulating_assets}, is directly used to perform path planning. This means that the $C(n_p,n)$ values of each expanded node are measured by running the dynamic simulator and by measuring the driving energy with the method in \ref{sub:energy}. As this approach has access to the ground truth energy value during planning, it is able to provide exact predictions and the solution with the lowest driving energy, with a predicted and real energy value of \SI{15.623}{kJ}. Therefore, the solution provided by Conv1D requires \SI{9.5}{\percent} more driving energy than the ideal ground truth. However, DynSim assumes to have full domain knowledge of the environment during planning. While this is possible in the case of a simulator, it is often impractical in real-world scenarios. In contrast, Conv1D does not require any specific domain knowledge to be implemented, as it learns how to make energy estimations by collecting data. Hence, while in this paper we only focus on simulation, the Conv1D methodology can be in principle equally applicable to simulated and real-world scenarios. Moreover, the computational time of DynSim is measured to be more than \num{20} times higher than the other two methods. This provides insights on the often prohibitive computational workload of dynamic simulators for real-time planning. Specifically, we observe that, while Conv1D and RampModel have similar average node expansion times at around \SI{0.2}{s}, DynSim takes on average \SI{5.6}{s}. The three methods are tested on an Intel Core i9-9940X CPU and, for the feed forward computation of the neural network, on a GeForce RTX 2080 Ti GPU.

\begin{figure}[t]
		\centering
		\includegraphics[scale=0.134]{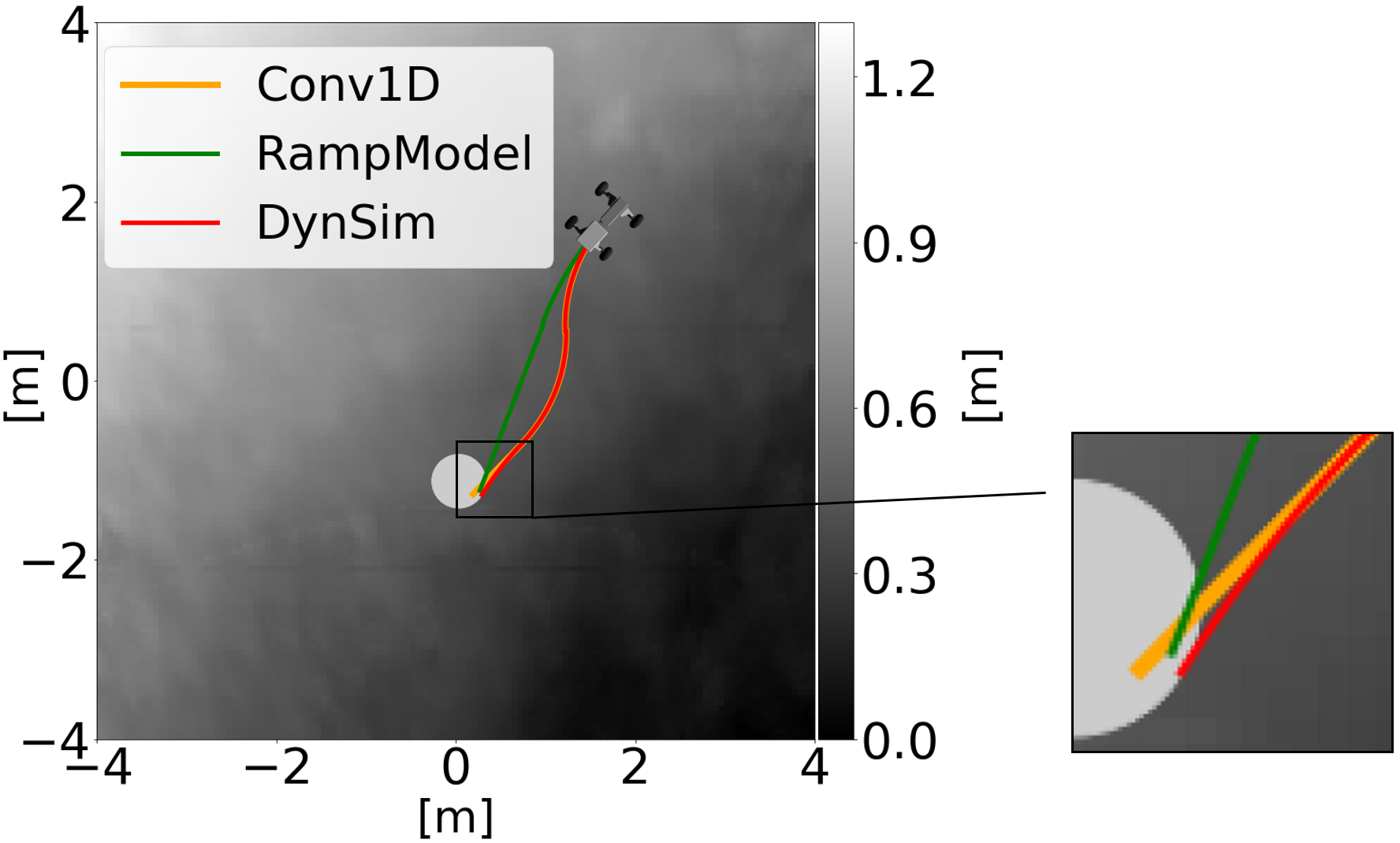}
		\caption{Three different solutions to the A* optimisation with different energy estimation methods.}
		\label{fig:path_planning_example}
	\end{figure}

	An example of the three different solutions according to the three energy estimation methods is given in Fig. \ref{fig:path_planning_example}, while their quantitative results are shown in the second part of Table \ref{tab:path_planner_test_results}. As with the previous results, DynSim has an exact prediction accuracy, and provides the best solution with predicted and real energy values of \SI{27.72}{J}. However, it also has the highest computational time with a total planning time of \SI{51.8}{s}. In contrast, Conv1D provides a slightly more energy demanding solution with predicted and real energy value of respectively \SI{29.23}{J} and \SI{30.44}{J}, but in a fraction of the total planning time, with a value of \SI{3.0}{s}. Moreover, the trajectory provided by Conv1D almost entirely corresponds to the one of DynSim. This provides evidence of the estimation accuracy of Conv1D, which results in a similar solution to the one selected by the ground truth method. Finally, RampModel has the shortest planning time, but also the lowest prediction accuracy, and the most energy demanding solution. We can interestingly observe how RampModel estimates its solution to provide a small quantity of energy recovery equal to \SI{-0.99}{J}. The robot proceeds on a slightly downhill trajectory and with a moderate roll inclination, which would lead to a marginal source of energy recovery if the surface was planar. \fontdimen2\font=2.3pt However, the presence of unstructured geometries increases the actual energy to a value of \SI{39.33}{J}. This confirms the higher robustness of our method which, by considering the actual terrain geometry, can provide more informed estimations and more energy efficient paths.
	
	\section{Conclusion and Future Works}\label{sec:conclusion}
	\fontdimen2\font=2.4pt
	In this study, an integrated energy-aware prediction and planning framework is presented for mobile robots in unstructured environments. We remark the benefit of our methodology to (1) perform accurate driving energy estimations over unstructured terrain geometries by capturing the sequential context of the traversed terrain, and (2) reduce the robot driving energy by predicting and planning directly over feasible trajectories. We demonstrate the advantages of our methodology with respect to an inclination-based heuristic energy model to increase the prediction r2 score over planned trajectories by \SI{66.8}{\percent} and reduce the driving energy consumption by \SI{5.5}{\percent}. The considerable improvement of prediction accuracy is of particular relevance, as in many robotic applications an incorrect estimation of energy can compromise the safety and success of the mission.
	
	We are extending this work in several directions. While in the current implementation trajectories were traversed at constant speed, the ability of our architecture to process data sequentially holds great promises for analysing the effect on driving energy of variable velocities. In addition, while this study analysed the effect on driving energy of terrain geometry, future works will consider strategies to include different terrain properties as an additional input to the proposed architecture. Finally, we intend to conduct experimental tests on a real robotic platform \cite{marcel} to validate the results obtained during simulation.

	\section*{Acknowledgement}
	\fontdimen2\font=3pt This work has been carried out within the framework of the EUROfusion Consortium and has received funding from the Euratom research and training programme under grant agreement No 633053. The views and opinions expressed herein do not necessarily reflect those of the European Commission. The authors are very grateful to the Autonomous Systems Group of RAL SPACE for providing the SEEKER dataset.

	
	
	\bibliographystyle{IEEEtran}
	\bibliography{IEEEfull,bibliography}
	
\end{document}